\documentclass{article}
\usepackage[preprint,nonatbib]{neurips_2022}
\usepackage[utf8]{inputenc} % allow utf-8 input
\usepackage[T1]{fontenc}    % use 8-bit T1 fonts
\usepackage{hyperref}       % hyperlinks
\usepackage{url}            % simple URL typesetting
\usepackage{booktabs}       % professional-quality tables
\usepackage{nicefrac}       % compact symbols for 1/2, etc.
\usepackage{microtype}      % microtypography
\usepackage{xcolor}         % colors
\usepackage{cite}
\usepackage{amsmath,amssymb,amsfonts}
\usepackage{amsthm}
\usepackage{algorithm}
\usepackage[noend]{algpseudocode}
\usepackage{graphicx}
\usepackage{textcomp}
\usepackage{xcolor}

\newcommand{\BEA}{\begin{eqnarray}}
\newcommand{\EEA}{\end{eqnarray}}
\newcommand{\argmin}{\mathop{\mathrm{argmin}}}
\newcommand{\argmax}{\mathop{\mathrm{argmax}}}

\def \RR{\mathbb{R}} 
\def \A {\mathcal{A}} %%% Architecture set function
\def \E {\mathbf{E}} %%% Expectation function
\def \P {\mathcal{P}} %%% Scenario defining Metric
\def \L {\mathcal{L}} %%% Loss function
\def \D {\mathcal{D}} %%% data set

\title{Transfer-Once-For-All: AI Model Optimization \\ for Edge}
\author{%
Achintya Kundu \\
IBM Research \\
\texttt{achintya.k@ibm.com} \\
\And
Laura Wynter \\
IBM Research \\
\texttt{lwynter@sg.ibm.com} \\
\AND
Rhui Dih Lee \\
IBM Research \\
\texttt{rhui.dih.lee@ibm.com} \\
\And
Luis Angel Bathen \\
IBM Research \\
\texttt{bathen@us.ibm.com} \\
}

\begin{document}

\maketitle

\begin{abstract}
Weight-sharing neural architecture search aims to optimize a configurable neural network model (supernet) for a variety of deployment scenarios across many devices with different resource constraints. Existing approaches use evolutionary search to extract models of different sizes from a supernet trained on a very large data set, and then fine-tune the extracted models on the typically small, real-world data set of interest. The computational cost of training thus grows linearly with the number of different model deployment scenarios. Hence, we propose Transfer-Once-For-All (TOFA) for supernet-style training on small data sets with constant computational training cost over any number of edge deployment scenarios. Given a task, TOFA obtains custom neural networks, both the topology and the weights, optimized for any number of edge deployment scenarios.  To overcome the challenges arising from small data, TOFA utilizes a unified semi-supervised training loss to simultaneously train all subnets within the supernet, coupled with on-the-fly architecture selection at deployment time. 
\end{abstract}

%%%=============================
\section{Introduction}
%%%=============================
Recent work in the field of weight-sharing neural architecture search (NAS) has proposed to jointly train a huge number of models of different sizes, and hence many resource constraint levels, at once \cite{ofa_mit,compOFA,attentiveNAS}. This class of methods aims to optimize a configurable neural network model,  referred to as a ``supernet",  to address deployment requirements across devices with different resource constraints. The supernet training paradigm can thus provide optimized models for a wide variety of edge hardware. Indeed,  weight-sharing NAS has become the dominant approach to finding specialized neural networks for edge scenarios \cite{ofa_mit}.

Supernet training has been shown to perform well on large data sets, such as in \cite{alphaNet}. When it comes to inference on specific tasks,  existing supernet approaches use evolutionary search, as proposed by \cite{ofa_mit}, to extract a number of individual models of different sizes based on floating point operations (Flops) or inference time (also called latency) from the supernet trained on the large data set.   Then, those individual extracted models, i.e., subnetworks, or ``subnets", are fine-tuned on the typically small, real-world data set of interest. The computational cost of fine-tuning for each task thus grows linearly with the number of different model deployment scenarios. For inference on the edge, it is common to have a large number of deployment scenarios. As such, existing approaches are very costly in practice.

Hence, we propose transfer-once-for-all (TOFA) for supernet-style training on small data sets with constant computational training cost over any number of edge deployment scenarios. This is accomplished by training the TOFA supernet on the task data itself. A key idea of TOFA is to leverage an existing neural network and efficiently transform it into a task-specific supernet, whilst simultaneously searching for architectures that offer the best trade-off across the deployment objectives of interest. Therefore, unlike existing supernet-based NAS, we combine supernet transfer learning with the search process. At the conclusion of this process, TOFA returns subnets that span the entire objective trade-off front and are already trained to the task. The subnets produced by TOFA can thus be utilized without additional training or fine-tuning. 

However, the challenge created by the TOFA approach is that the supernet must be trained on what is typically a data set of much smaller size than would be the case when training a supernet on a large data set used in typical pre-training. To overcome the challenges arising from small data, TOFA thus utilizes a  pre-trained neural network as the starting point to obtain the best of both worlds: high accuracy from large-data pre-training, and computational cost reduction through supernet training on the task itself. To combat the problem of supernet training on small data sets, TOFA employs a unified semi-supervised training loss to simultaneously train all subnets within the supernet, coupled with on-the-fly architecture selection at deployment time. 

In this paper, we address the following questions and make the following contributions: 
\begin{itemize}
    \item First, is it possible to bypass the pre-training of a weight-sharing supernet and perform weight-sharing training directly on the task data set of interest? Is there a means to benefit from the pre-training phase indirectly? In this paper, we provide an affirmative answer to both of these questions. We illustrate the benefit of directly training a supernet on small, e.g. transfer learning, data sets, whilst still leveraging the weights obtained from large data set pre-training. Our experimental results show that the accuracies of individual models extracted from a TOFA supernet trained on the task itself are better or as good as those from individually fine-tuned models extracted from a pre-trained supernet, without the computational cost. TOFA supernet training offers constant computational cost as opposed to the linearly growing training cost involved in fine-tuning individual models for each edge deployment scenario.
    
    \item Second, in existing approaches, appropriate subnetwork models,  henceforth called ``subnets", are extracted from a supernet through an evolutionary search process. The evolutionary search requires first training an accuracy predictor and then using it in the search or else measuring the actual validation accuracy for each subnet. Thus, an evolutionary search procedure is computationally very costly. In addition, on real-world, end-user tasks, extracting a validation set is often not possible given the very small set of labelled data available for training. Is it possible to define an alternative to evolutionary search to save computation time and still maintain performance?
    We answer this question also in the affirmative. We observe that the variation of accuracy across subnets within a given size range (e.g. in number of parameters) is relatively small. As such we propose a zero-cost approach that selects subnets from the training process with no additional computational costs. We show that a fixed selection rule such as selecting the most recently sampled subnet works remarkably well in practice. 
    
    \item Third, can supernet training converge effectively on small data sets that are typical in real-world tasks? Again, the answer is yes. Our solution is to train supernets not only on the labelled training data used for typical task fine-tuning, but to add unlabelled data also into the supernet training. In this paper, we thus propose a novel loss function to train the supernet in a semi-supervised manner. Experimentally we show better performance on small real-world data sets.
\end{itemize}

In the next section, we provide the necessary background and discuss related work. Section 3 presents the TOFA approach. In Section 4, we provide experimental results with TOFA, and we conclude with interesting avenues for future work in Section 5.

%%%--------------------------------------------------
\begin{figure*}[hbt!]
	\centering
	\includegraphics[width=.9\textwidth]{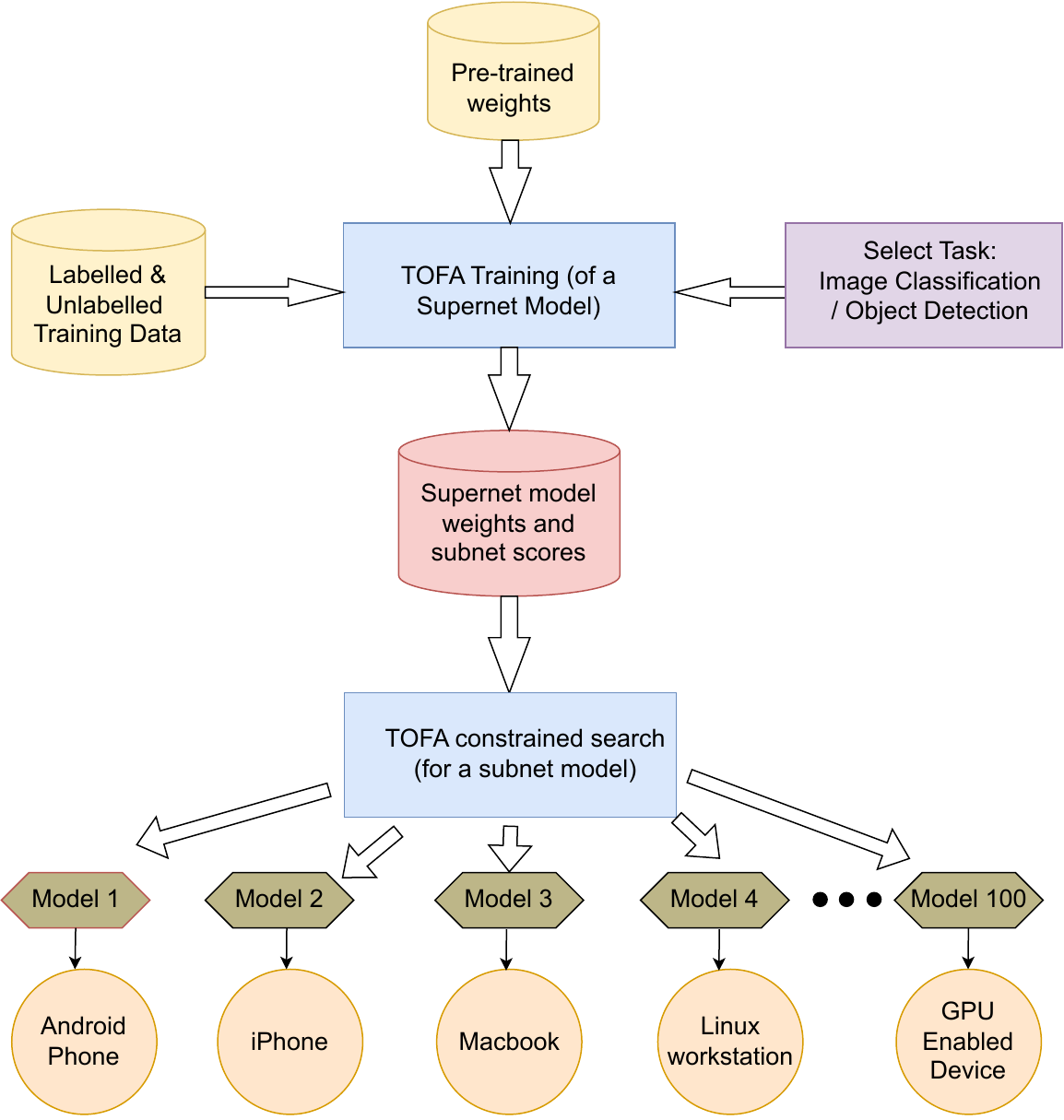}
	\caption{The TOFA framework: One-shot training for many edge-device specific model deployment scenarios}\label{fig:tofa}
\end{figure*}
%%%--------------------------------------------------

%%%====================================
\section{Background and Related Work}
%%%====================================

%%%--------------------------------------------------
\subsection{ Weight-Sharing Neural Architecture Search}
%%%--------------------------------------------------
Existing weight-sharing NAS methods employ a two-stage approach to neural network training for subsequent inference on the edge. In the first stage, a weight-sharing supernet is trained. Then, in the second stage candidate subnetworks with weights inherited directly from the supernet are sampled. This approach was appealing as it provided an efficient alternative to individually training each different subnet architecture when multiple deployment scenarios for edge inference require different model sizes.  

The benefit of this approach can be seen to be due to the ability to share weights during training across networks of different sizes, thus gaining computational efficiency. See, for example \cite{kshotNAS_Neurips21, trainingHeuristicsWeightSharingNAS , weightSharingRandomSearch_CVPR20,ofa_mit, dong2022priorguided}.
The seminal paper  \cite{ofa_mit} introduced a progressive shrinkage approach for training their so-called Once-For-All (OFA) network in a cascading manner. Later, CompOFA \cite{compOFA} proposed reducing the search space of OFA to bring down the training complexity.  On the other hand, the Dynamic-OFA approach \cite{dynamicOFA}  extends the OFA approach to address the need to dynamically change the model size, but the authors only consider this problem on very large data sets such as those used in pre-training.  

An alternative to the multi-stage progressive shrinkage style of training a supernet is to train all subnets simultaneously from the beginning of training. This joint approach has gained popularity and one might say has become the dominant approach for supernet training. See for example BigNAS \cite{bigNAS}, AttentiveNAS \cite{attentiveNAS}, AlphaNet \cite{alphaNet}, FastOFA \cite{fastOFA}.
In \cite{neural_arch_transfer}, neural architecture transfer (NAT) was proposed to fine-tune a pre-trained supernet on a downstream task so that a palette of models can be extracted. That work involves a prohibitively costly loop involving multiple rounds of full supernet training based on an empirical distribution, accuracy prediction, evolutionary search, accuracy measurement on a validation set, and then updating the empirical distribution and re-running the full supernet training again, and so on. 

Weight-sharing NAS however comes with its own set of challenges such as accuracy benefits to some subnetworks but accuracy degradation in others. We revisit this in the next section where we present the TOFA method.

%%%--------------------------------------------------
\subsection{Semi-Supervised Learning (SSL) and FixMatch}
%%%--------------------------------------------------
Because TOFA is concerned with producing a palette of models of different sizes for direct use on end-user tasks, TOFA must be able to perform supernet training effectively on small data sets.
We thus propose to leverage unlabelled data in addition to the small labelled data through a novel loss function to train the supernet in a semi-supervised manner. 

FixMatch \cite{FixMatch} proposes a  semi-supervised training loss that combines two approaches to semi-supervised learning (SSL): consistency regularization and pseudo-labeling. FixMatch combines these two techniques through the use of separate weak and strong data augmentation. Consistency regularization attempts to utilize unlabelled data by relying on the assumption that the model should output similar predictions when fed perturbed versions of the same image. On the other hand, pseudo-labeling leverages training with a labelled loss where the model itself is used to obtain the "guessed" labels for unlabelled data. The FixMatch loss term is particularly well-suited to TOFA supernet training as it contains the two key components -- consistency regularization and pseudo labelling -- yet the overhead in calculating the loss term is negligible. Other semi-supervised learning methods such as \cite{openmatch, SSL_eccv, MixMatch} require significant additional overhead in the form of costly data augmentation or training a separate binary classifier to select the unlabelled data, with marginal benefits compared to FixMatch on the supernet training process.

%%%--------------------------------------------------
\subsection{CNN Image Classification Models}
%%%--------------------------------------------------
In this paper, we consider the image classification problem as an exemplar of TOFA. Image classification is a fundamental task in computer vision, where, given a data set and possibly multiple objectives to optimize, one seeks to learn a model to classify images. The most prevalent convolutional neural network (CNN) models used for image classification include ResNet \cite{ResNet}, MobileNet \cite{MobileNetv3}, EfficientNet \cite{EfficientNet}, FBNet \cite{FBNetv3}. We are particularly interested in MobileNet as it is well-suited to use on edge devices.

%%%====================================
\section{ TOFA Method }
%%%====================================

%%%--------------------------------------------------
\subsection{The TOFA Algorithm}
%%%--------------------------------------------------
Instead of training every architecture sampled during NAS search from scratch, procedures using NAS with weight sharing inherit weights from previously trained networks or from a supernet. Directly inheriting weights obviates the need to optimize the weights from scratch and can speed up the search from thousands of GPU days to only a few. The key to the success of weight sharing is that the performance of the subnets with the inherited weights be highly correlated with the performance of the same subnet when thoroughly trained from scratch. This suggests that the supernet weights should be learnt in such a way that all subnets are optimized simultaneously. We achieve this by sampling subnet architectures during supernet training itself, where the probabilities of training any particular subnet for each batch of data is uniform in expectation. To regularize the supernet training, we  use a ``sandwich rule'', that is, largest and smallest subnets are  sampled along with two random subnets at every iteration. 

Since we allow for integration of unlabelled data in the TOFA training phase, we denote the labelled training data set by $\D_{l}:=\{ (x_i^l,y_i^l)\}_{i=1}^{N_l}$ and the unlabelled training data by $\D_{u}:=\{ x^u_i\}_{i=1}^{N_u}$. A labelled validation/test data set is used to evaluate learning performance of a model. The goal of TOFA is to train a supernet model $\theta$ by solving 
\BEA
 \argmin_{\theta} ~\E_{a \in \A }\,[\L(\theta_a; \D_{tr})\,]\,,%\label{eq:ofa_obj}
\EEA
where $\mathcal{L}$  denotes the loss of a subnet $a$ on a training data set, $\D_{tr}$,  and the expectation is over allowed subnets $a \in \A$. In this paper, we thus propose a novel way of training a supernet to reap benefits from  additional unlabelled training data when available.

When no unlabelled data is available, i.e., in the fully supervised setting, one can replace the loss on the training data $\D_{tr}$ with that of the labelled training data $\D_{l}$. Here, we formulate the supernet training as the following optimization problem:
\BEA
 \argmin_{\theta} ~\E_{a \in \A }[\,\L(\theta_a; \D_{l},\D_{u})\,],%\label{eq:ofa_obj}
\EEA
where $\mathcal{L}$  denotes the loss function of a subnet $\theta_a$ on both the labelled as well as unlabelled training data sets $\D_{l}, \D_{u}$ and the expectation is over allowed sub-models $a \in \A$. This immediately raises the question of finding a suitable loss function $\L$ that works not only for labelled examples but is also applicable for unlabelled samples. To address this question, we propose novel loss functions later in this section. Before that, we describe below the basic set-up for our supernet training algorithm.

For a given image classification task, we thus focus on simultaneously learning a palette of convolutional neural network (CNN) models of different computational complexities that can be readily deployed across edge devices with varying resource (that is, inference time / memory) constraints. To do so, TOFA involves learning a weight-sharing CNN supernet  where the CNN's architectural dimensions such as kernel size, depth, number of channels, expansion ratio, etc. can be set dynamically from a fixed finite set. A subnet model is defined as a static model obtained by setting each of the dynamic architectural dimensions of the supernet model. The specification of a supernet includes weights $\theta$ of the network with the maximum configuration for each architectural dimension, called the ``maxnet",  and an architecture search space $\A$ consisting of all possible subnets. As the weights of subnet models are inherited  from an appropriate subset of the supernet, we denote the weights of a particular subnet configuration $a \in \A$ by $\theta_a$ to emphasize the fact that weights $W$ of the supernet are shared across all sub-nets. The TOFA supernet training algorithms is described in Algorithm \ref{alg:TOFA_algo}.

\begin{algorithm}[hbt!]
	\caption{TOFA Supernet Training} 
	\label{alg:TOFA_algo}
	\begin{algorithmic}[1]
	\State \textbf{Require:} A supernet with  architecture space $\A$ and maxnet weights $\theta$. Strong and weak data augmentation functions: $s[\cdot], \,w[\cdot]$. Number of iterations $T$.
    \State \textbf{Initialisation:} Use pre-trained weights to initialize $\theta$. 
    \State The list of sampled subnets $\Omega \,=\,\{\}$.
	\For{t = 1, \ldots, T} 
        \State Get a mini-batch of $B_l$ of labelled samples.\
		\State \emph{optimizer.zero\textunderscore grad()}. 
        \State \emph{loss} ~=~ $\displaystyle \sum_{(x_l, \,y_l) \in B_l}CE( \theta_{max}(w[x_l]),y_l) $ 
        \State Get a mini-batch of $B_u$ unlabelled samples. 
        \State Predict labels: $\hat{y}_u:= \argmax\, \theta_{max}(w[x_u]) \, \forall x_u \in B_u$. 
        \State Set $\delta_u$ to $1$ if $\theta_{max}(w[x_u]) > 0.95$ and zero otherwise.
        \State \emph{loss} ~+=~ $\displaystyle \sum_{x_u \in B_u}\delta_u CE( \theta_{max}(s[x_u]),\hat{y}_u)$
        \State  Randomly sample architectures $A_{r_1}, A_{r_2}$ from $\A$.
       % \State Construct $\A_{sample}:= \{ A_{r_1}, A_{r_2}, A_{min}\}$.
        \For{$A ~in~\A_{sample} := \{ A_{r_1}, A_{r_2}, A_{min}\} $} 
        \State \emph{loss} ~+=~ $\displaystyle \sum_{(x_l,\, y_l) \in B_l} CE( \theta_{A}(w[x_l]),\theta_{max}(w[x_l]))$
        \State \emph{loss} ~+=~ $\displaystyle \sum_{x_u \in B_u} CE( \theta_{A}(w[x_u]),\theta_{max}(w[x_u]))$
        \EndFor
        \State \emph{loss.backward()}. 
        \State \emph{optimizer.step()}. 
        \State Update iteration score of the sampled subnets as $\forall A \in \A_{sample}$:~$Iteration(A) ~=~ t$.
       % \[ Iteration(A) ~=~ t ,~\mbox{if}~A \notin \Omega,\]
      %  \[ Iteration(A) ~+\!=~ t ,~\mbox{if}~A \in \Omega.\]
        \State Update the list of sampled subnets: $\Omega ~\leftarrow~ \Omega \cup \A_{sample}$.
	\EndFor
	\end{algorithmic}
\end{algorithm}

\begin{table*}
    \centering
    \caption{An illustration of the TOFA search space.}
    \label{tab:space}
    \begin{tabular}{c|ccccc}
    \toprule % from booktabs package
    \bfseries Block & 
    \bfseries Width & 
    \bfseries Depth & 
    \bfseries Kernel size & 
    \bfseries Expansion ratio & 
    \bfseries Squeeze \& Excitation \\
    \midrule % from booktabs package
Conv & \{16,\,24\} & - & 3 & - & - \\

MBConv-1 & \{16,\,24\} & \{1,\,2\} & \{3,\,5\} & 1 & No \\
MBConv-2 & \{24,\,32\} & \{3,\,4,\,5\} & \{3,\,5\} & \{2,\,4,\,6\} & No \\
MBConv-3 & \{32,\,40\} & \{3,\,4,\,5,\,6\} & \{3,\,5\} & \{2,\,4,\,6\} & Yes \\
MBConv-4 & \{64,\,72\} & \{3,\,4,\,5,\,6\} & \{3,\,5\} & \{2,\,4,\,6\} & No \\
MBConv-5 & \{112,\,128\} & \{3,\,4,\,5,\,6,\,7,\,8\} & \{3,\,5\} & \{2,\,4,\,6\} & Yes \\
MBConv-6 & \{192,\,216\} & \{3,\,4,\,5,\,6,\,7,\,8\} & \{3,\,5\} & \{2,\,4,\,6\} & Yes \\
MBConv-7 & \{216,\,224\} & \{1,\,2\} & \{3,\,5\} & \{2,\,4,\,6\} & Yes \\
MBPool & \{1792,\,1984\} & - & 1 & 6 & - \\
    \bottomrule % from booktabs package
    \end{tabular}
\end{table*}

%%%====================================
\subsection{Loss Function}
%%%====================================
Consider an image classification problem with $C$-classes. Given an image input $x$, we denote the model output for a  particular subnet architecture $r$ by $\theta_{r}(x)$. %$ \in \Delta^C$, where $\Delta^C:= \{ (z_1, \cdots, z_C)\,|\, \sum_{j=1}^Cz_j=1,\,z_i \ge 0, \forall i \}$. 
We employ two different data augmentation techniques: a weak augmentation $w[\cdot]$ for both labelled as well as unlabelled data, and a strong augmentation $s[\cdot]$ only for the unlabelled data as required by the FixMatch loss. We denote a labelled sample by $(x_l, y_l)$ and unlabelled sample by $x_u$. For an input sample $x$ we associate a ``guessed'' label as $\hat{y} := \argmax \theta_{max}( w[x])$  (the class label predicted by the maxnet). We consider the following loss functions for the  labelled and unlabelled samples:
\BEA
L^{lab}_{max}(x_l,y_l) &:=& CE( \theta_{max}( w[x_l] ), \, LS_{\alpha}(y_l) ), \nonumber \\
L^{unlab}_{max}(x_u) &:=& CE( \theta_{max}( s[x_u]), \,LS_{\alpha}(\hat{y}_u)  ),  \nonumber \\
L^{dist}_{sub}(x) &:=& \sum_{i=1}^2 CE( \theta_{A_{{r_i}}}( w[x]), \, \theta_{max}( w[x] ) )  \nonumber \\
& & + \,CE( \theta_{min}( w[x]), \, \theta_{max}( w[x] ) ) ,
\nonumber 
\EEA
where $LS_{\alpha}$ stands for Label-Smoothing with parameter $\alpha$, $CE(\cdot,\cdot)$ is the standard Cross Entropy loss and $A_{r_1}, A_{r_2}$ are two randomly selected subnet configurations.

When there is no unlabelled data, we propose the following loss function for use in TOFA training:
\BEA
\L^{lab}(\theta;B_{l}) := \sum_{(x_l,y_l)\in B_{l}} \,[\, L^{lab}_{max}(x_l,y_l) \,+ \,
%\sum_{(x_l,y_l)\in B_{l}}  
L^{dist}_{sub}(x_l)\,]\,,
\EEA
With unlabelled data as well as labelled data available, we have two  choices for the loss function corresponding to the unlabelled data. Specifically,
\BEA
\L^{FM}(\theta;B_{u}) &:=& \sum_{x_u \in B_{u} \,:\, \max( \theta_{max}( w[x_u])) > 0.95} L^{unlab}_{max}(x_u) ,\nonumber \\
\L^{dist}(\theta;B_{u}) &:=& \sum_{x_u \in B_{u} } L^{dist}_{sub}(x_u) ,\nonumber 
\EEA
where $\L^{FM}$ is the FixMatch loss \cite{FixMatch}, $B_l$ is a mini-batch of labelled training data, $B_u$ denotes a mini-batch of unlabelled data, and $\L^{dist}$ is the distillation loss term.

In the ablation studies performed, we explore the following potential TOFA loss functions for the setting which includes both labelled and unlabelled training data:
\BEA
\L^{lab}_{TOFA}(\theta) \!&:=& \! \E [\L^{lab}(\theta;B_{l}) ] , \\
\L^{FM}_{TOFA}(\theta)\!&:=& \! \E [\L^{lab}(\theta;B_{l})  + \L^{FM}(\theta;B_{u})] ,  \nonumber \\
\L^{dist}_{TOFA}(\theta)\!&:=& \! \E [\L^{lab}(\theta;B_{l})  + \L^{dist}(\theta;B_{u})],  \nonumber \\
\L_{TOFA}(\theta) \!&:=& \! \E [\L^{lab}(\theta;B_{l}) \! + \! \L^{FM}(\theta;B_{u}) \!+ \! \L^{dist}(\theta;B_{u})], \nonumber  \label{eq:semiloss} 
\EEA
where $\E[\cdot]$ denotes expectation over the randomly sampled mini-batches $B_l, B_u$ and random subnets $A_{r_1}, A_{r_2}$ used in the distillation loss, $L^{dist}_{sub}(\cdot)$. Note that at one end $\L^{lab}_{TOFA}$ only relies on the labelled training data whereas $\L_{TOFA}$ utilizes all three loss terms: the loss on the labelled data, the FM loss on the unlabelled data and the distillation loss on the unlabelled data. In the experimental section, we perform an ablation study on performance of TOFA with the above variants. In Algorithm ~\ref{alg:TOFA_algo}, we present TOFA with $\L_{TOFA}$ as it is the optimal choice of loss function in TOFA supernet training.

%%%=============================
\subsection{TOFA Search Space}
%%%=============================
Searching for optimal network architectures can be performed over different model configuration spaces. The versatility of the chosen search space has a direct impact on the quality of the models obtained from the search procedure. We adopt a modular design for each subnet, consisting of a head, multiple stages and a tail. The head and tail are common to all subnets and are not searched. For the searchable part, we segment the CNN architecture into seven sequentially connected stages. The stages gradually reduce the feature map size and increase the number of channels. Each stage comprises of multiple layers where each layer has an inverted residual bottleneck structure. Specifically, for the CNN architectures and the edge scenarios of interest, a MobileNetv3 block is employed. 

For the search space design, we closely follow FBNetV3. In particular, while we use the same structure as FBNetV3, we reduce the search range of channel widths, depths and expansion ratios. We also limit the largest possible subnet in the search space to be less than 2500 MFlops and constrain the smallest subnet to be larger than 200 Mflops. This provides significantly greater efficiency and better performance. We search for the input image resolution $h \times h$ from $ h \in \{192, \, 224,\, 256,\, 288,\, 320\}$. In each stage, we search over the number of layers, where only the first layer uses stride 2 if the feature map size decreases, and we allow each block to have minimum of two and maximum of eight layers. In each layer, we search over the expansion ratio $e$ between the number of output and input channels of the first $1 \times 1$ convolution and the kernel size of the depth-wise separable convolution. We use a fixed $k \in \{3,\,5\}$ to decide the kernel size $k \times k $ for all layers in a particular stage. The expansion ratio $e$ is selected from $\{2,4,6\}$ and same $e$ is used across all layers in a stage. In summary, we search over the five primary hyperparameters of CNNs i.e., the depth $d$, that is, the number of layers in a stage, the width $c$, that is, the number of output channels, the kernel size $k$, the expansion ration $e$, and the input image resolution $h$. %The resulting search space consists of approximately $10^{19}$ subnet configurations. 

%%%=============================
\subsection{Resource-Constrained Searching}
%%%=============================
In addition to high predictive accuracy, end user applications demand NAS algorithms to simultaneously balance several conflicting objectives that are specific to each deployment scenario. For instance, mobile or edge devices have restrictions in terms of model size, Flops (floating point operations, i.e., number of multiply-adds), latency, power consumption, and peak memory footprint. With no prior assumption on the correlation among these objectives, a scalable subnet selection algorithm is required to drive the search towards the high dimensional Pareto efficiency front of such a multi-objective function. 

One of the  contributions of TOFA is the observation that it is often unnecessary to use  a costly evolutionary search as a second phase in weight-sharing NAS. In addition, given the typically very small data sets available in real-world, end-user tasks, training an accuracy predictor or even extracting a validation set is often not possible in practice. TOFA includes a fixed subnet selection rule to select a subnet within each size range. The pseudo-code is provided in Algorithm \ref{alg:subnet_search} and works as follows:  initialize the index of all subnets to zero. Every time a subnet is sampled during supernet training, say at iteration $k$, we update the index of the sampled subnet by adding the iteration number $k$ to its previous index. At the end of the training phase, we store the set of subnets that have non-zero indices. We call this set $\Omega$ and for any subnet $a \in \Omega$ we denote its index by $s(a)$. Let $p:\A \to \RR$ denote a desired model selection metric such as flops/ inference time on a specific device. Then, given a budget $t$,  the proposed subnet selection strategy can be expressed as:
\BEA
\argmax_{ a \in \Omega } ~ s(a)  \,\,\,\mbox{s.t.} \,\,\, p(a) \le t. \nonumber
\EEA
Note that in the training phase, TOFA jointly optimizes all possible subnet architectures specified in the search space through weight sharing without considering resource constraints.

The Fixed subnet selection rule works  well for edge resource constraints that scale proportionally to the number of parameters. 
In edge scenarios with more complex resource constraints such as constraints on multiple metrics or single metrics that do not scale proportionally to number of parameters, and where there is sufficient data to extract a validation set, a two-stage  approach as in other NAS works can be used as follows.  Once TOFA training is completed,  one solves:
\BEA
\{ A_i \} ~= & \argmax_{A_i \in \A} \P(\theta^{*}_{A_i}; \D_{val}) \nonumber \\
& ~\mbox{s.t.}~~ \mbox{Flops}(A_i)\,\le\, c_i, ~i=1,\ldots,m,
\label{eq:constrain_search}
\EEA
where $\P(\cdot,\D_{val})$ measures predictive performance (say accuracy) of a subnet model $A_i$ on the validation set $\D_{val}$, $\theta^{*}$ denotes the trained supernet weights, and Flops constraints $\{ c_i \}_{i=1}^m$ are selected to cover the Flop range spanned by the supernet. Naturally, \eqref{eq:constrain_search} can be generalized to other metrics and can be executed on-the-fly without retraining or fine-tuning the supernet.

\begin{algorithm}[hbt!]
	\caption{Fixed subnet selection rule (last sampled) } 
	\label{alg:subnet_search}
	\begin{algorithmic}[1]
    	\State \textbf{Input:} $\Omega$: list of subnets sampled during training along with their $Iteration$ at which it was sampled, $p$: performance metric (flop), $t$: budget constraint 
     \State \textbf{Compute performance metrics:} $p(A)$ for all $A \in \Omega$.
    \State Set $\Omega_t ~:= ~\{ A \in \Omega \,: \, p(A) \le t\}$ 
    \State Find $A_t^* = \argmax_{ A \in \Omega_t } Iteration(A)$
    \State \textbf{Output:} Subnet configuration $A_t^*$.
 \end{algorithmic}
\end{algorithm}

%%-------------------------------------
\subsection{Comparison of Training Complexity}
%%--------------------------------------
In this section, we compare the computational cost of training TOFA with that of training a single model. To illustrate the advantage of training a supernet, we assume that there are $N$ different inference use cases, each demanding a model satisfying a specific requirement. In case of single model training, the total computational cost involved in training grows linearly with the number of models ($N \times$ the cost of training one model). On the contrary, TOFA incurs one-time fixed training cost irrespective of the number of different models.

%%%=============================
\section{Experimental Results}
%%%=============================
In this section, we present experimental results to evaluate the efficacy of TOFA on multiple image classification tasks. Object detection can also benefit from TOFA although the semi-supervised loss term is less relevant on object detection tasks. 
In addition, we perform an ablation study on the use of different loss functions in the  setting where both labelled and unlabelled data are used in TOFA training.

%%-------------------------
\subsection{Data Sets}
%%-------------------------
We consider six image classification data sets for evaluation with sample size varying from $2,040$ to $100,000$ images (20 to 6,000 images per class). These data sets span a wide variety of image classification tasks from high-level recognition (CIFAR-10, CIFAR-100) to fine-grained recognition  (Stanford Cars, FGVC Aircraft, Oxford-IIIT Pets, Oxford Flowers102) which represent a realistic range of end user tasks \cite{EfficientNet, alphaNet}. We use the ImageNet data set for pre-training the supernet and the same pre-trained weights are used for transfer learning on all the above data sets. For each of the transfer learning datasets, Table~\ref{tab:datasets} shows the number of total training samples as well as the number of labelled training samples used in the semi-supervised experiments. The labelled training samples are randomly selected from the total training set as $3$ samples/class (for Flower dataset) to $20$ samples/class (for Cifar-100 dataset). The remaining training samples are treated as unlabelled samples in our semi-supervised experiments. 

\begin{table}
    \centering
    \caption{Training Data set sizes}\label{tab:datasets}
    \begin{tabular}{lccc}
      \toprule % from booktabs package
      \bfseries Data set & 
       \bfseries Classes &
      \bfseries Training Samples &
      \bfseries  Labelled Samples \\ 
      \midrule % from booktabs package
      \midrule
      Aircraft  &  100 & 6667 & 1300 \\
      Car       &  196 & 8144  & 2156\\
      Flower    &  102 & 2040  & 306 \\
      Pet       &  37  & 3680   & 259\\
      Cifar-10  &  10  & 50000  & 200\\
      Cifar-100 &  100 & 50000  & 2000\\
      \bottomrule % from booktabs package
    \end{tabular}
\end{table}

%%-------------------------
\subsection{Implementation Details}
%%-------------------------
\paragraph{Data Augmentation} For weak augmentation we perform random color jitter, random horizontal flip and random size crop \& resize. For stronger data augmentation, we use ``AutoAugment" function from torchvision library in addition to the weak augmentation techniques. We do not perform any hyper-parameter tuning for TOFA. We mention the default values of the hyper-parameters below.
\paragraph{Mini batch size} For all experiments we use a mini-batch size of 32 for labelled data as well as unlabelled data.
\paragraph{Learning Rate} We use a default learning rate of $0.01$ in a cosine learning rate scheduler with $5$ warm up epochs.
\paragraph{Regularization} We use dropout probability of $0.3$ and 
drop connect rate of $0.2$ for the fully connected layer in the classifier head. We also use the squared $L_2$ norm regularization (weight decay) $0.00001$ for all model weights except bias terms.
\paragraph{Optimizer} We use standard stochastic gradient descent (SGD) as optimizer with a maximum of $10,000$ iterations.
\paragraph{ Label Smoothing} In TOFA, while computing cross entropy loss with one-hot labels, we employ label smoothing with coefficient $0.1$.

%%%%-------------------------------------------
\subsection{TOFA Performance using Labelled and Unlabelled Data}
%%-----------------------------------------------
\begin{table}
    \centering
    \caption{Performance comparison of TOFA maxnet using labelled and unlabelled data against the alternatives}\label{tab:data}
    \begin{tabular}{lcccc}
      \toprule % from booktabs package
      \bfseries data set & 
      \bfseries Baseline &
      \bfseries  TOFA  & \bfseries TOFA  & 
      \bfseries Baseline \\
\bfseries  & 
      \bfseries  labelled & 
      \bfseries   labelled & \bfseries lab+unlab & 
      \bfseries  lab+unlab \\ 
      \midrule % from booktabs package
      \midrule
      Aircraft  & 65.6 & 69.0 &\textbf{ 75.2 }& 73.4 \\
    %            & Minnet &  & 61.7  & 71.2 &  \\ \midrule
      Car  & 66.9 & 71.5 & \textbf{ 82.4 }& 80.3 \\
   %       & Minnet &  & 63.6 & 78.9 &  \\ \midrule
      Flower  & 80.6 & 81.7 & \textbf{ 92.3 }& 89.4 \\
    %            & Minnet &  & 79.8 & 92.0 &  \\ \midrule
      Pet  & 81.8 & 83.0 & \textbf{87.6 }& 87.4 \\
   %             & Minnet &  & 78.3 & 86.2 &  \\ \midrule
     Cifar-10  & 70.4 & 72.3 & \textbf{83.5 }& 82.6 \\
    %           & Minnet &  & 66.9 & 80.9 &  \\ \midrule
      Cifar-100  & 71.0 & 73.6 & \textbf{79.0 }& 77.6 \\
    %            & Minnet &  & 67.2 & 75.4 &  \\
      \bottomrule % from booktabs package
    \end{tabular}
\end{table}
In this section we evaluate the benefit accrued by adding unlabelled data in the TOFA supernet training procedure. To do so, we compare the  performance of the TOFA maxnet trained using both labelled and unlabelled data against (a) TOFA trained with  only labelled data,  (b)  single model training on only labelled data, denoted as the Baseline, and (c)  single model training   using both labelled and unlabelled data with the FixMatch \cite{FixMatch}  method, denoted FixMatch. The results are presented in Table \ref{tab:data}. We initialize all models with the same pre-trained weights as in TOFA. 

In a fully-supervised setting with no unlabelled data, the TOFA maxnet outperforms the Baseline which also uses only labelled data by margin of 2-4$\%$. This shows an  advantage of TOFA's supernet based training which reaps benefit from training different model architectures together. In the setting where both labelled and unlabelled data are available,  TOFA further improves the accuracy across data sets and  beats the stand alone semi-supervised model trained using   FixMatch.

%%%---------------------------------------------
\subsection{Performance Comparison with Alternative Approaches using only Labelled Data}
%%%---------------------------------------------
In this section we compare TOFA performance with that of other approaches. In order to compare all approaches on equal footing, 
we assume in this section that there is access to only a labelled training data set with no additional unlabelled data,
as  other approaches do not allow for incorporation of unlabelled data. Specifically, here, the full training set with  labels is used to train all models and performance accuracy is evaluated on the full test set. We use the default training/test splits from TorchVision library for these standard data sets. From Table \ref{tab:data}, we  saw that TOFA as defined for both labelled and unlabelled data outperforms TOFA on only labelled data on all end user data sets. 

\begin{figure}[hbt]
	\centering
	\includegraphics[width=.75\textwidth]{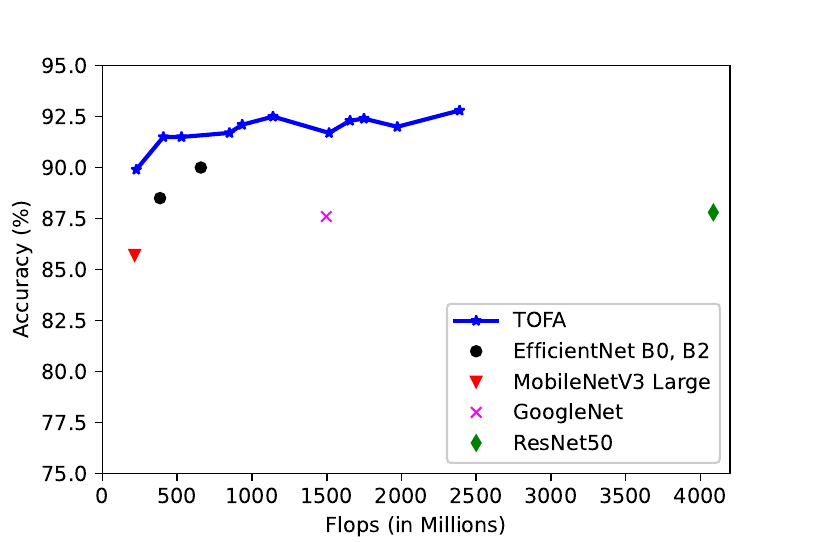}
	\caption{Performance Comparison on Stanford Cars data set between TOFA using only labelled data, two small networks developed for inference on edge devices (MobileNet and EfficientNet) and two large networks (GoogLeNet and ResNet50). TOFA offers better performance across a wide range of resource levels (in Flops).}\label{fig:Car}
\end{figure}

\begin{figure}[hbt]
	\centering
	\includegraphics[width=.8\textwidth]{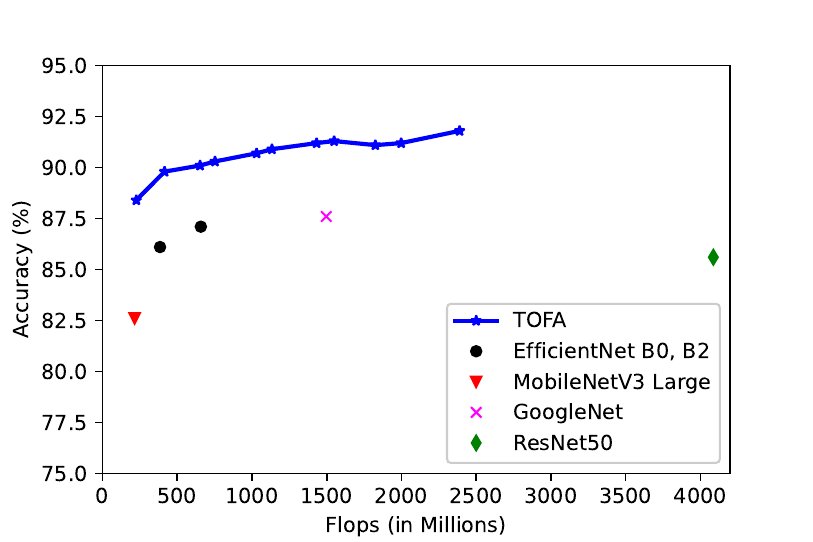}
	\caption{Performance Comparison on FGVC Aircraft data set between TOFA using only labelled data, two small networks developed for inference on edge devices (MobileNet and EfficientNet) and two large networks (GoogLeNet and ResNet50). TOFA offers better performance across a wide range of resource levels (in Flops).}\label{fig:Aircraft}
\end{figure}

\begin{figure}[hbt]
	\centering
	\includegraphics[width=.8\textwidth]{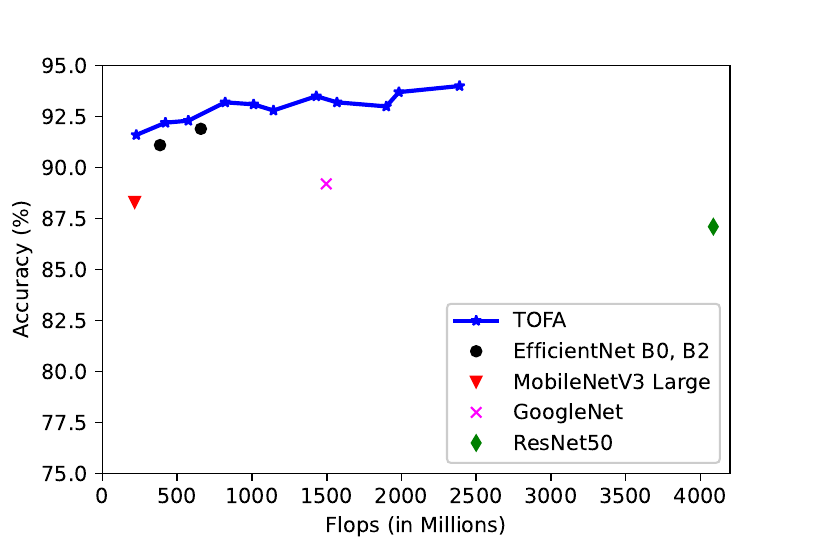}
	\caption{Performance Comparison on Oxford Pets dataset between TOFA using only labelled data, two small networks developed for inference on edge devices (MobileNet and EfficientNet) and two large networks (GoogLeNet and ResNet50). TOFA offers better performance across a wide range of resource levels (in Flops).}\label{fig:Pet}
\end{figure}

Here, in Figures ~\ref{fig:Car}, \ref{fig:Aircraft} and \ref{fig:Pet},  we compare the accuracy of TOFA with only labelled data  against other approaches.   Model complexity is measured in terms of number of Flops for a single forward pass.   TOFA supernet training  produces a palette of trained models covering a range of Flops from $\sim 200$ Million  to $\sim 2400$ Million. For all  three standard data sets: Stanford Cars (Fig.~\ref{fig:Car}), FGVC Aircrafts (Fig.~\ref{fig:Aircraft}) and Oxford Pets (Fig.~\ref{fig:Pet}), the subnet models produced by one-shot TOFA trained on only labelled data   outperforms individually fine-tuned state-of-the-art image classification models such as Mobilnetv3 Large, Resnet50, GoogLeNet as well as EfficientNet B0 \& B2 under the given Flops budgets. These Flops budgets are suitable for deployment on resource-constrained edge devices.

\begin{figure}[hbt]
	\centering
	\includegraphics[width=.8\textwidth]{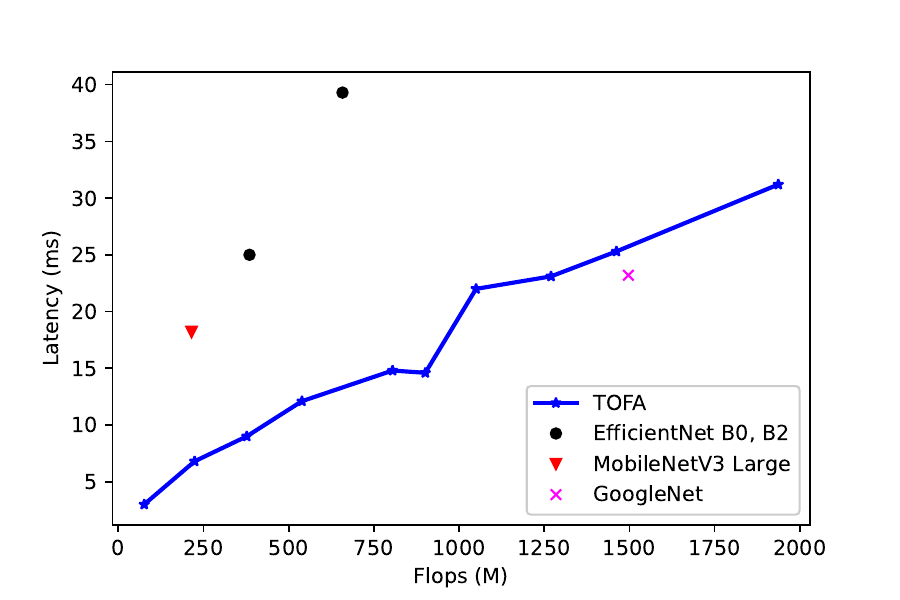}
	\caption{Comparison of Latency on MacbookPro Laptop between TOFA using only labelled data, two small networks developed for inference on edge devices (MobileNet and EfficientNet) and a large network (GoogLeNet).   TOFA offers lower latency than all the other models at the lower end of the resource consumption levels. Only GoogLeNet has a slightly lower latency than the larger TOFA models.}\label{fig:flop_lat}
\end{figure}

\begin{figure}[hbt]
	\centering
	\includegraphics[width=.8\textwidth]{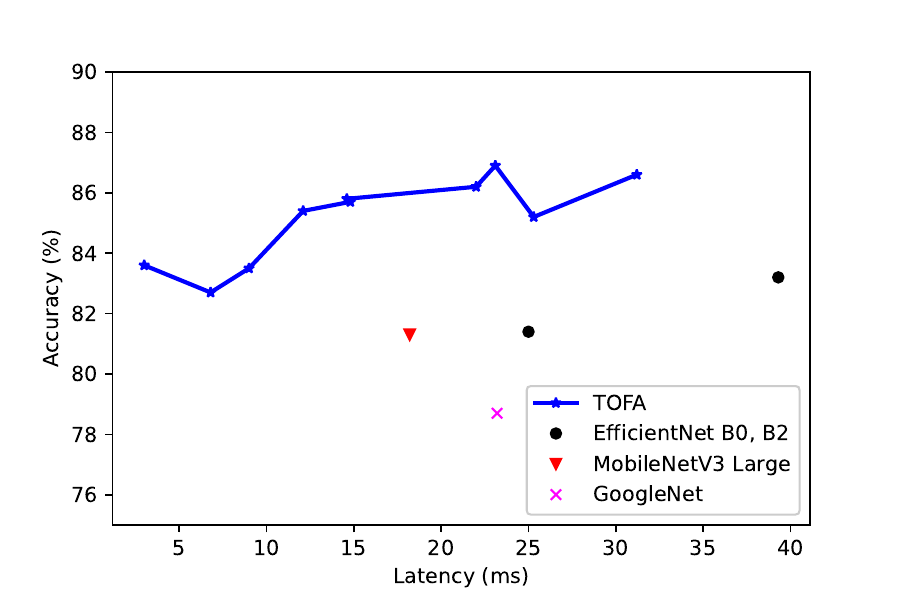}
	\caption{Latency vs Accuracy Trade-off (on CIFAR-100 dataset) of different models: TOFA using only labelled data, two small networks developed for inference on edge devices (MobileNetv3 and EfficientNet) and a large network (GoogLeNet). TOFA offers better accuracy across all latency levels.}\label{fig:lat_acc}
\end{figure}

Next we compare TOFA using only labelled data against alternative approaches  in terms of inference time, i.e., latency. Figure \ref{fig:flop_lat} provides a summary of the latency comparison across the range of TOFA subnet models as compared to MobileNetv3 (Large) and EfficientNet (B0 \& B2) which are specially designed CNN architectures suitable for inference on edge devices, and against   GoogLeNet. ResNet50 requires more than 4000  mega Flops with a corresponding high inference time, yet provides poor accuracy.  Hence we exclude ResNet50 from Figure \ref{fig:flop_lat} so as to focus on the region of interest, which is models requiring less than 2000 mega Flops. TOFA latency is far lower than the alternatives with the exception of GoogLeNet at the higher end of the Flops range.

Similarly, we examine the Latency vs Accuracy Trade-off (on CIFAR-100 dataset) in Figure \ref{fig:lat_acc}  between TOFA using only labelled data,  MobileNet and EfficientNet,  and GoogLeNet. TOFA even with only labelled data  offers clearly better accuracy across all latency levels.

%%%---------------------------------------------
\subsection{Device specific Latency and Memory Measurement}
\begin{figure}[hbt]
	\centering
	\includegraphics[width=.8\textwidth]{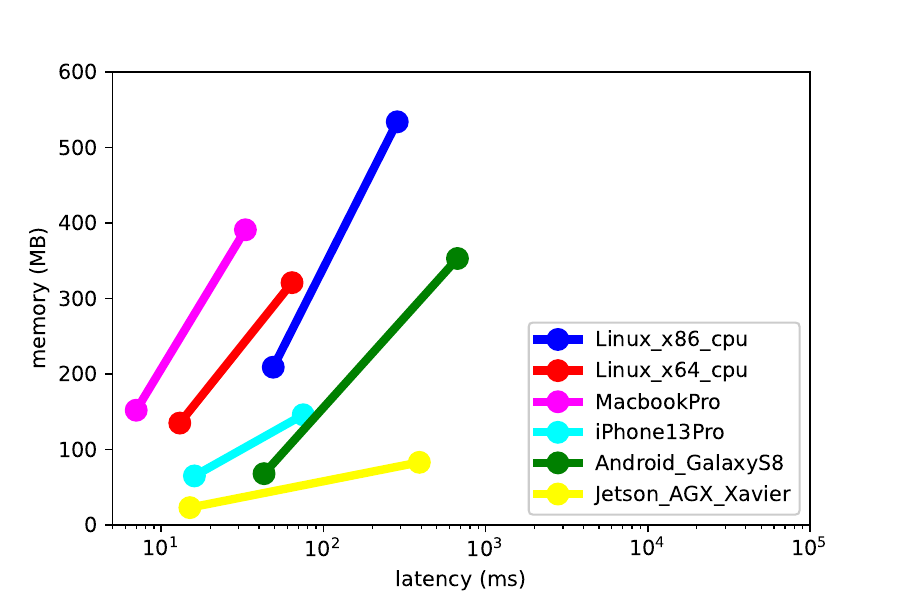}
	\caption{Device specific Latency and Memory usage of TOFA models. TOFA subnetworks can be used for inference across a wide range of devices.}\label{fig:lat_mem}
\end{figure}

%%%---------------------------------------------
In this section, we measure the actual inference time (latency) and peak memory usage of the TOFA subnet models on several different edge devices: Linux x86 cpu based workstation, Jetson AGX Xavier, Android Samsung Galaxy phone, Macbook Pro laptop as well as iPhone13Pro. Among these devices only Jetson AGX Xavier utilizes computation on GPU (others rely only on CPU) while running inference. We run inference test on the two extreme subnet configurations (the maxnet, which is the largest subnet model in the supernet, and the minnet, which is the smallest subnet model in the supernet) on each of these devices to compute the range of latency / peak memory for the TOFA models. In Figure \ref{fig:lat_mem}, we see that TOFA models can be deployed on a wide range of devices with different latency \& memory budgets.  

%%%---------------------------------------------
\subsection{Ablation Study on Fixed Subnet  Selection Rule}
\begin{figure}[hbt]
	\centering
	\includegraphics[width=.7\textwidth]{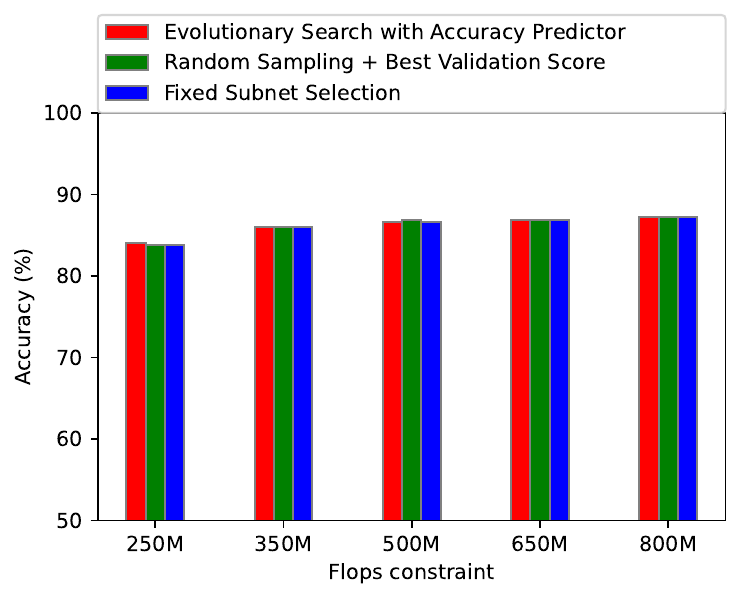}
	\caption{Performance of different subnet selection schemes on CIFAR-100 Dataset. Our Fixed subnet selection rule  offers comparable performance to the much costlier subnet extraction methods.}\label{fig:subnet_cifar100}
\end{figure}

\begin{figure}[hbt]
	\centering
	\includegraphics[width=.7\textwidth]{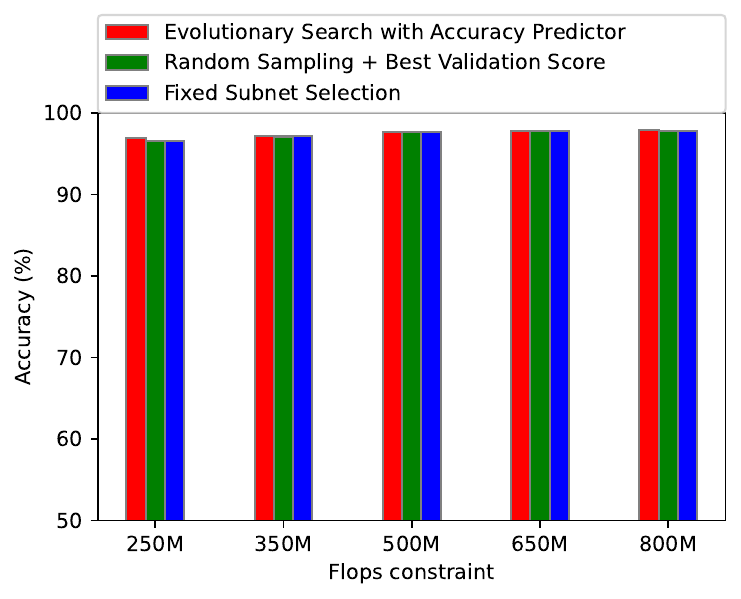}
	\caption{Performance of different subnet selection schemes on CIFAR-10 Dataset. Our Fixed subnet selection rule offers comparable performance to the much costlier subnet extraction methods.}\label{fig:subnet_cifar10}
\end{figure}

\begin{figure}[hbt]
	\centering
	\includegraphics[width=.7\textwidth]{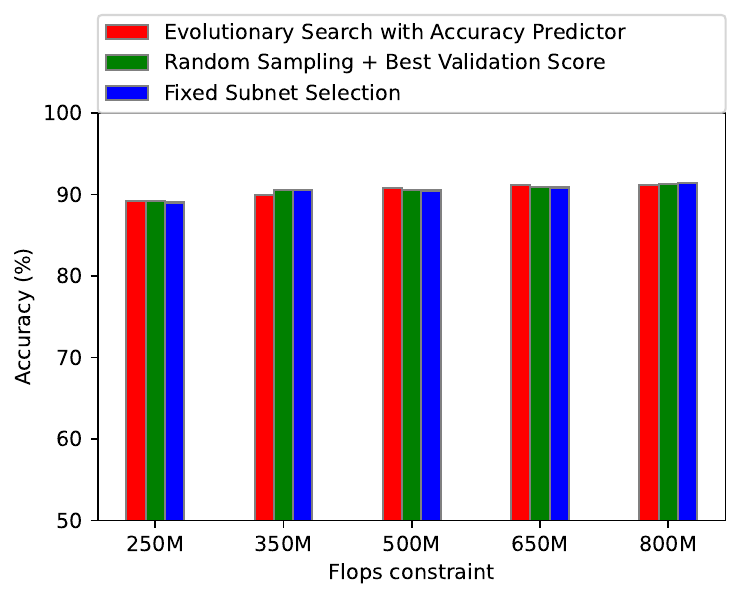}
	\caption{Performance of different subnet selection schemes on Stanford Cars Dataset. Our Fixed subnet selection rule offers comparable performance to the much costlier subnet extraction methods.}\label{fig:subnet_car}
\end{figure}

\begin{figure}[hbt]
	\centering
	\includegraphics[width=.7\textwidth]{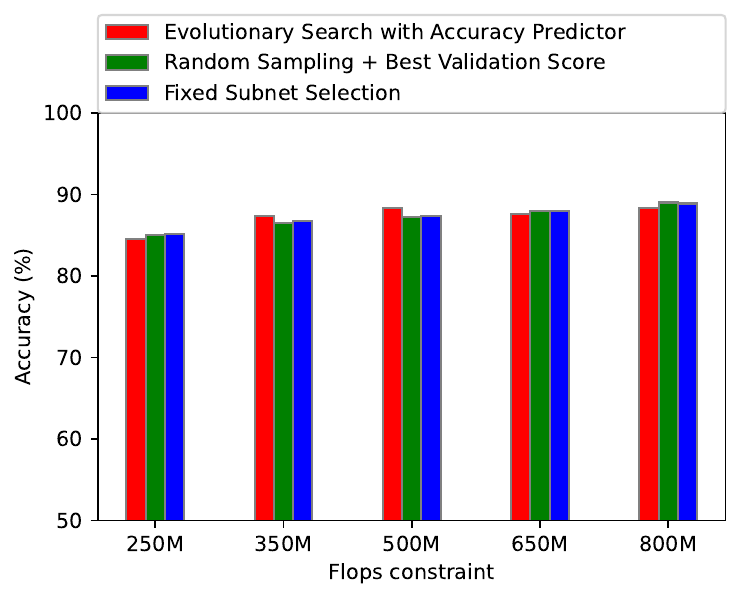}
	\caption{Performance of different subnet selection schemes on FGVC Aircrafts Dataset. Our Fixed subnet selection rule offers comparable performance to the much costlier subnet extraction methods.}\label{fig:subnet_aircraft}
\end{figure}

%%%---------------------------------------------
In this section, we  compare experimentally the effectiveness of different subnet selection schemes given a Flops constraint. In Figures \ref{fig:subnet_cifar100}, \ref{fig:subnet_cifar10}, \ref{fig:subnet_car} and \ref{fig:subnet_aircraft}, we compare the performance of the Fixed subnet selection rule with that of (a) an accuracy predictor based on Evolutionary search and (b) random sampling with best accuracy determined  on a validation set. For approach (a) and (b), we randomly sampled total 1000 subnet configurations spread across the flops range and evaluated accuracy on the full validation set. Here, the 30\% of the full  training dataset is used as a validation set and rest 70\% is used for training the supernet.

We can observe  that there is no significant difference in accuracy across these subnet selection methods. On the other hand, the computational complexity of the evolutionary search and the best validation score-based schemes are extremely high and require the availability of a validation set. Evolutionary search, for example,  requires an accuracy predictor model, which in turn requires a one-time computation of validation accuracy of $\sim\!\!1000$ different subnets.  Random sampling requires computing the validation accuracy of $\sim\!\!10$ different subnets for every Flops constraint search. The Fixed subnet selection rule, on the other hand, needs neither validation accuracy computation nor any accuracy predictor and as such is essentially zero-cost. Because the variance across subnets of similar size is small enough, this zero-cost method works remarkably well, as shown in Figures \ref{fig:subnet_cifar100}- \ref{fig:subnet_aircraft}.

\begin{table}
    \centering
    \caption{Ablation study on effects of semi-supervised loss terms in TOFA}\label{tab:data2}
    \begin{tabular}{llcccc}
      \toprule % from booktabs package
      \bfseries Data set & 
      \bfseries Model & 
      \bfseries $\L^{lab}_{TOFA}$ & 
      \bfseries $\L^{FM}_{TOFA}$ &
     \bfseries $\L^{dist}_{TOFA}$ &
     \bfseries $\L_{TOFA}$ \\
      \midrule % from booktabs package
      \midrule
      Aircraft & Maxnet & 69.0 &  71.5 & 73.8 & \textbf{ 75.2}\\
                & Minnet & 61.7 & 65.1 & 70.0 & \textbf{71.2 }\\ \midrule
      Car & Maxnet &  71.5 &  78.4 & 79.9 & \textbf{ 82.4 }\\
          & Minnet &  63.6 &  72.6 & 75.5 & \textbf{78.9}\\ \midrule
      Flower & Maxnet & 81.7 & 92.0 & 91.5 & \textbf{92.3}\\
             & Minnet & 79.8 &  90.9 & 91.2 & \textbf{92.0}\\ \midrule
                
      Pet & Maxnet & 83.0 &  87.1 & 85.6 & \textbf{87.6  }\\
            & Minnet & 78.3 &  83.5 & 84.4 &  \textbf{86.2}\\ \midrule

     Cifar-10 & Maxnet & 72.3 &  82.7 & 80.5 & \textbf{83.5}\\
               & Minnet & 66.9 &  76.0 & 79.1 & \textbf{80.9 }\\ \midrule

      Cifar-100 & Maxnet & 73.6 &  77.6 & 76.9 & \textbf{79.0 }\\
                & Minnet & 67.2 &  72.1 & 73.1 & \textbf{75.4 }\\
      \bottomrule % from booktabs package
    \end{tabular}
\end{table}

%%%---------------------------------------------
\subsection{Ablation Study on Semi-supervised Loss Functions}
%%%---------------------------------------------
Finally, we investigate the effect of different loss terms in \eqref{eq:semiloss} for training TOFA using both labelled and unlabelled data. The results are shown in Table  \ref{tab:data2}. We report performances of the maxnet (the largest subnet model in the supernet) as well as the minnet (the smallest subnet model in the supernet). The first observation is that all of our loss function variants   perform better than the fully-supervised setting ($\L^{lab}_{TOFA}$) in which no unlabelled data is used. Secondly, observe that adding the distillation loss on the unlabelled samples, even without using the pseudo-label based loss, already provides a significant positive impact on  performance. The best results can be seen in the last column which incorporates both unlabelled loss terms in addition to the labelled loss, which is  the default loss function $\L_{TOFA}$ used in Algorithm \ref{alg:TOFA_algo}.

%%==================================
\section{Conclusions and Discussion}
%%===================================
We have introduced TOFA: Transfer-Once-For-All, as a means to reap the benefits of supernet style NAS training for end user tasks on edge devices. We have shown that it is possible to perform weight-sharing supernet NAS on end user small data sets by the incorporation of an initialization of the maxnet from a  pre-trained network and the use of both labelled and unlabelled data. In addition, through an observation that the variance across subnet performance in each size range is generally very low, we introduced a zero-cost alternative to  the much more costly subnet extraction approaches using a Fixed subnet selection rule. 

We illustrate in this paper the application of TOFA to CNN architectures for image classification. Our current research  involves adapting TOFA  to transformer architectures  for use on large language models.

%%==================================
\bibliography{references_ofa}

\begin{thebibliography}{10}

\bibitem{weightSharingRandomSearch_CVPR20}
Gabriel Bender, Hanxiao Liu, Bo~Chen, Grace Chu, Shuyang Cheng, Pieter-Jan
  Kindermans, and Quoc~V. Le.
\newblock Can weight sharing outperform random architecture search? an
  investigation with {TuNAS}.
\newblock In {\em Proceedings of the IEEE/CVF Conference on Computer Vision and
  Pattern Recognition (CVPR)}, June 2020.

\bibitem{MixMatch}
David Berthelot, Nicholas Carlini, Ian Goodfellow, Nicolas Papernot, Avital
  Oliver, and Colin~A Raffel.
\newblock Mixmatch: A holistic approach to semi-supervised learning.
\newblock In H.~Wallach, H.~Larochelle, A.~Beygelzimer, F.~d\textquotesingle
  Alch\'{e}-Buc, E.~Fox, and R.~Garnett, editors, {\em Advances in Neural
  Information Processing Systems}, volume~32. Curran Associates, Inc., 2019.

\bibitem{ofa_mit}
Han Cai, Chuang Gan, Tianzhe Wang, Zhekai Zhang, and Song Han.
\newblock Once for all: Train one network and specialize it for efficient
  deployment.
\newblock In {\em International Conference on Learning Representations}, 2020.

\bibitem{FBNetv3}
Xiaoliang Dai, Alvin Wan, Peizhao Zhang, Bichen Wu, Zijian He, Zhen Wei, Kan
  Chen, Yuandong Tian, Matthew Yu, Peter Vajda, and Joseph~E. Gonzalez.
\newblock Fbnetv3: Joint architecture-recipe search using predictor
  pretraining, 2020.

\bibitem{dong2022priorguided}
Peijie Dong, Xin Niu, Lujun Li, Linzhen Xie, Wenbin Zou, Tian Ye, Zimian Wei,
  and Hengyue Pan.
\newblock Prior-guided one-shot neural architecture search.
\newblock 2022.

\bibitem{fastOFA}
Jun Fang, Li~Yang, Chengyao Shen, Hamzah Abdel-Aziz, David Thorsley, and Joseph
  Hassoun.
\newblock Fast and efficient once-for-all networks for diverse hardware
  deployment, 2022.

\bibitem{ResNet}
Kaiming He, Xiangyu Zhang, Shaoqing Ren, and Jian Sun.
\newblock Deep residual learning for image recognition.
\newblock {\em CoRR}, abs/1512.03385, 2015.

\bibitem{MobileNetv3}
Andrew Howard, Mark Sandler, Grace Chu, Liang{-}Chieh Chen, Bo~Chen, Mingxing
  Tan, Weijun Wang, Yukun Zhu, Ruoming Pang, Vijay Vasudevan, Quoc~V. Le, and
  Hartwig Adam.
\newblock Searching for mobilenetv3.
\newblock {\em CoRR}, abs/1905.02244, 2019.

\bibitem{dynamicOFA}
Wei Lou, Lei Xun, Amin Sabet, Jia Bi, Jonathon Hare, and Geoff~V. Merrett.
\newblock Dynamic-ofa: Runtime dnn architecture switching for performance
  scaling on heterogeneous embedded platforms.
\newblock In {\em 2021 IEEE/CVF Conference on Computer Vision and Pattern
  Recognition Workshops (CVPRW)}, pages 3104--3112, 2021.

\bibitem{neural_arch_transfer}
Z.~Lu, G.~Sreekumar, E.~Goodman, W.~Banzhaf, K.~Deb, and V.~Boddeti.
\newblock Neural architecture transfer.
\newblock {\em IEEE Transactions on Pattern Analysis and Machine Intelligence},
  43(09):2971--2989, 2021.

\bibitem{compOFA}
Manas Sahni, Shreya Varshini, Alind Khare, and Alexey Tumanov.
\newblock {C}omp{OFA}: Compound once-for-all networks for faster multi-platform
  deployment.
\newblock In {\em Proc. of the 9th International Conference on Learning
  Representations}, 2021.

\bibitem{openmatch}
Kuniaki Saito, Donghyun Kim, and Kate Saenko.
\newblock Openmatch: Open-set semi-supervised learning with open-set
  consistency regularization.
\newblock In A.~Beygelzimer, Y.~Dauphin, P.~Liang, and J.~Wortman Vaughan,
  editors, {\em Advances in Neural Information Processing Systems}, 2021.

\bibitem{FixMatch}
Kihyuk Sohn, David Berthelot, Nicholas Carlini, Zizhao Zhang, Han Zhang,
  Colin~A Raffel, Ekin~Dogus Cubuk, Alexey Kurakin, and Chun-Liang Li.
\newblock Fixmatch: Simplifying semi-supervised learning with consistency and
  confidence.
\newblock In H.~Larochelle, M.~Ranzato, R.~Hadsell, M.F. Balcan, and H.~Lin,
  editors, {\em Advances in Neural Information Processing Systems}, volume~33,
  pages 596--608. Curran Associates, Inc., 2020.

\bibitem{kshotNAS_Neurips21}
Xiu Su, Shan You, Mingkai Zheng, Fei Wang, Chen Qian, Changshui Zhang, and
  Chang Xu.
\newblock K-shot nas: Learnable weight-sharing for nas with k-shot supernets.
\newblock In Marina Meila and Tong Zhang, editors, {\em Proceedings of the 38th
  International Conference on Machine Learning}, volume 139 of {\em Proceedings
  of Machine Learning Research}, pages 9880--9890. PMLR, 18--24 Jul 2021.

\bibitem{EfficientNet}
Mingxing Tan and Quoc Le.
\newblock {E}fficient{N}et: Rethinking model scaling for convolutional neural
  networks.
\newblock In Kamalika Chaudhuri and Ruslan Salakhutdinov, editors, {\em
  Proceedings of the 36th International Conference on Machine Learning},
  volume~97 of {\em Proceedings of Machine Learning Research}, pages
  6105--6114. PMLR, 09--15 Jun 2019.

\bibitem{alphaNet}
Dilin Wang, Chengyue Gong, Meng Li, Qiang Liu, and Vikas Chandra.
\newblock Alphanet: Improved training of supernet with alpha-divergence.
\newblock {\em arXiv preprint arXiv:2102.07954}, 2021.

\bibitem{attentiveNAS}
Dilin Wang, Meng Li, Chengyue Gong, and Vikas Chandra.
\newblock Attentivenas: Improving neural architecture search via attentive
  sampling.
\newblock {\em arXiv preprint arXiv:2011.09011}, 2020.

\bibitem{bigNAS}
Jiahui Yu, Pengchong Jin, Hanxiao Liu, Gabriel Bender, Pieter{-}Jan Kindermans,
  Mingxing Tan, Thomas~S. Huang, Xiaodan Song, Ruoming Pang, and Quoc~V. Le.
\newblock Bignas: Scaling up neural architecture search with big single-stage
  models.
\newblock 2020.

\bibitem{trainingHeuristicsWeightSharingNAS}
Kaicheng Yu, Ren{\'{e}} Ranftl, and Mathieu Salzmann.
\newblock How to train your super-net: An analysis of training heuristics in
  weight-sharing {NAS}.
\newblock {\em CoRR}, abs/2003.04276, 2020.

\bibitem{SSL_eccv}
Qing Yu, Daiki Ikami, Go~Irie, and Kiyoharu Aizawa.
\newblock Multi-task curriculum framework for open-set semi-supervised
  learning.
\newblock In Andrea Vedaldi, Horst Bischof, Thomas Brox, and Jan-Michael Frahm,
  editors, {\em Computer Vision -- ECCV 2020}, pages 438--454, Cham, 2020.
  Springer International Publishing.

\end{thebibliography}
%%==================================
\end{document}